\documentclass[letterpaper, 10 pt, conference]{ieeeconf}  

\IEEEoverridecommandlockouts                              

\usepackage{caption}
\usepackage{mdframed}
\usepackage{changepage}
\usepackage{pifont}
\usepackage{hyperref}
\usepackage{amssymb}
\let\labelindent\relax

\usepackage{enumitem}

\newenvironment{noindentenum}
  {\begin{enumerate}[leftmargin=1.5em,labelsep=0.5em]}
  {\end{enumerate}}

\newenvironment{noindentitem}
  {\begin{itemize}[leftmargin=1em,labelsep=0.5em]}
  {\end{itemize}}

\title{\LARGE \bf
Mitigating Cross-Modal Distraction and Ensuring Geometric Feasibility \\via Affordance-Guided and Self-Consistent MLLMs for Task Planning \\in Instruction-Following Manipulation}

\author{
    Yu-Hong Shen$^{1\ddagger}$ \, Chuan-Yu Wu$^{2\ddagger}$ \, Yi-Ru Yang$^{1}$ \, Yen-Ling Tai$^{1}$ \, Yi-Ting Chen$^{1}$%
    \thanks{$^{\ddagger}$ Equal Contribution.}%
    \thanks{$^{1}$Department of Computer Science, National Yang Ming Chiao Tung University, Hsinchu City, Taiwan {\tt\small {stanley.shen2003.cs14,ben25000233.cs12,yling.cs12, ychen\}@nycu.edu.tw}
    }
    }%
    \thanks{$^{2}$Department of Computer Science and Information Engineering, National Taiwan University, Taipei City, Taiwan 
    {\tt\small r13922a23@ntu.edu.tw}
    }%
}

\begin{document}

\refstepcounter{figure}

\makeatletter
\let\@oldmaketitle\@maketitle
\renewcommand{\@maketitle}{\@oldmaketitle
    \includegraphics[width=\linewidth]{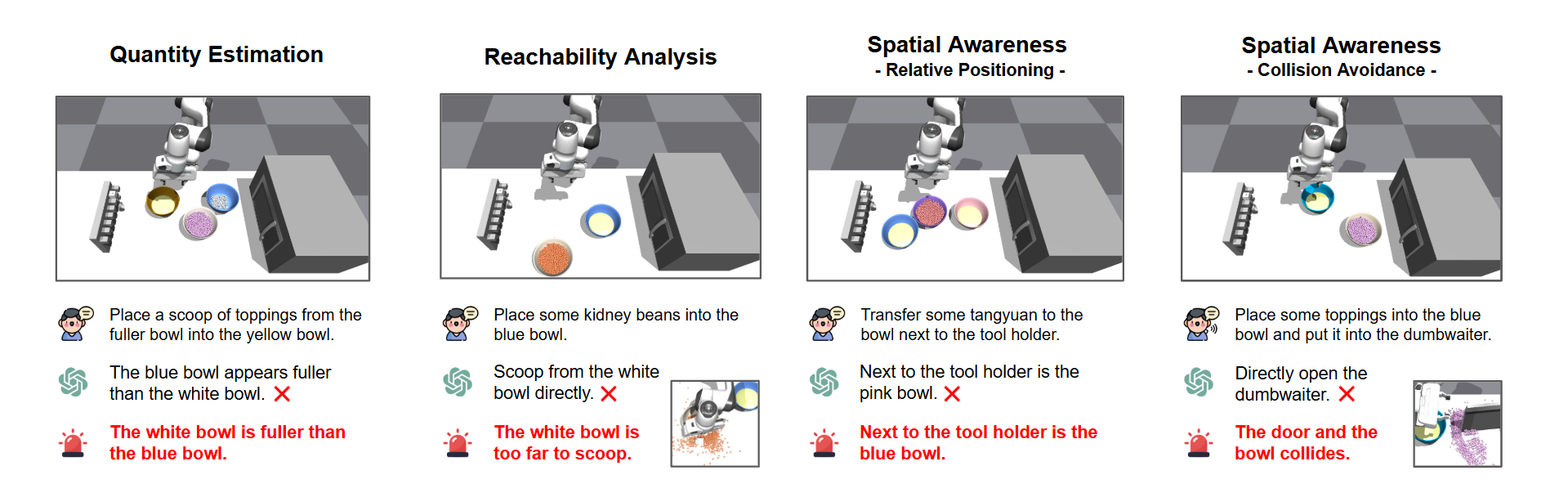}   
\noindent{\small\selectfont Fig.~\thefigure: Our benchmark illustrates four key requirements for closed-loop task planning in instruction-following manipulation: \textbf{quantity estimation}, \textbf{reachability analysis}, \textbf{relative positioning}, and \textbf{collision avoidance}. 
In the figures, each scenario, a customer issues a natural language command, e.g., “place a scoop of toppings from the fuller bowl into the yellow bowl.” An instruction-following agent must interpret the instruction, ground the referred items, perform the manipulation, and assess physical feasibility.
However, current multimodal large language models, when used as an instruction-following task planner, often struggle with cross-modal ambiguity and geometric constraints. 
Specifically, they fail at quantity comparison, object repositioning, spatial reasoning, and collision-free planning, resulting in incorrect or incomplete executions.}

    \label{fig-teaser}
    \bigskip}
\makeatother
\maketitle
\thispagestyle{empty}
\pagestyle{empty}

\begin{abstract}

We investigate the use of Multimodal Large Language Models (MLLMs) with in-context learning for closed-loop task planning in instruction-following manipulation.
We identify four essential requirements for successful task planning: quantity estimation, reachability analysis, relative positioning, and collision avoidance. 
However, existing benchmarks fail to support holistic evaluation across all these aspects.
To address this gap, we introduce \textbf{QuARC} (Quantity, Analysis, Relative positioning, Collision), a new benchmark based on a food preparation scenario that integrates all four challenges.
Using QuARC, we reveal two major limitations of current MLLMs: cross-modal distraction and geometric infeasibility. 
To tackle these, we adapt Chain-of-Thought with Self-Consistency to mitigate reasoning loss from cross-modal distractions and incorporate an affordance predictor to guide planning based on geometric feasibility.
Our comprehensive evaluation analyzes performance across multiple baselines and explains sources of improvement.
Our method achieves a 76.7\% success rate on the benchmark, significantly outperforming the ViLa baseline (36.7\%), without requiring additional finetuning.
Code and dataset are available at \href{https://hcis-lab.github.io/Affordance-Guided-Self-Consistent-MLLM}{https://hcis-lab.github.io/Affordance-Guided-Self-Consistent-MLLM}
\end{abstract}

\begin{table*}[t!]
    \centering
    \resizebox{\textwidth}{!}{%
    \begin{tabular}{lcccccc}
    \hline
        \textbf{Benchmark} & \textbf{Type} & \begin{tabular}{c}
    \textbf{Quantity} \\
    \textbf{Estimation}
    \end{tabular} &
    \begin{tabular}{c}
    \textbf{Reachability} \\
    \textbf{Analysis}
    \end{tabular} &
    \begin{tabular}{c}
    \textbf{Relative} \\
    \textbf{Positioning}
    \end{tabular} &
    \begin{tabular}{c}
    \textbf{Collision} \\
    \textbf{Avoidance}
    \end{tabular}\\

    \hline
        ALFRED \cite{ALFRED}        & Mobile Manipulation     & \ding{55} & \checkmark & \ding{55} & \ding{55} \\
        BEHAVIOR-1K \cite{li2023behavior}    & Mobile Manipulation     & \ding{55} & \checkmark & \checkmark & \ding{55} \\
        CALVIN \cite{calvin}     & Manipulation  & \ding{55} & \ding{55} & \checkmark & \ding{55} \\
        TableEnv \cite{Text2Motion}    & Manipulation  & \ding{55} & \checkmark & \ding{55} & \checkmark \\
    \hline
        \textbf{QuARC (Ours)}  & Manipulation  & \checkmark & \checkmark & \checkmark & \checkmark \\
    \hline
        \multicolumn{6}{r}{\textbf{\checkmark}: requirement supported \quad \ding{55}: not supported} \\
    \end{tabular}
    }
    \caption{Comparison of existing benchmarks across four essential requirements, i.e., \textbf{quantity estimation}, \textbf{reachability analysis}, \textbf{relative positioning}, and \textbf{collision avoidance}.}
    \label{tab:comparison}
\end{table*}
\section{INTRODUCTION}

Imagine entering a tofu pudding shop where an embodied agent prepares customized dessert orders. In front of the agent is a table filled with bowls of tofu pudding and various toppings. Customers interact using natural language, expressing preferences like, “place a scoop of toppings from the fuller bowl into the yellow bowl.”
To complete this long-horizon task, the agent must interpret the instruction, ground the specified food items, execute manipulation skills, evaluate action feasibility, and adapt its behavior accordingly. 
This demands a holistic understanding and reasoning, posing a significant challenge for instruction-following embodied AI \cite{Socratic, FlowPlan}.

We identify four essential requirements for successful task execution: \textbf{quantity estimation}, \textbf{reachability analysis}, \textbf{relative positioning}, and \textbf{collision avoidance}.
Given the inherent ambiguity of natural language, quantity estimation and relative positioning become particularly difficult when instructions are vague. For example, instructions “scoop from the fuller bowl” or “place the food into the bowl next to the tool holder” involve ambiguity. 
In such cases, the agent must semantically reason the instruction, perceive the environment, and ground referred objects across visual and linguistic modalities.
Additionally, reachability analysis ensures that the agent can position its end-effector within kinematic limits to interact with the target, while collision avoidance guarantees that planned motions do not cause unintended contact with surrounding objects.

To systematically study and evaluate these essential requirements, we propose QuARC (\textbf{Qu}antity, \textbf{A}nalysis, \textbf{R}elative positioning, \textbf{C}ollision), a new benchmark entered on a food preparation use case that integrates all four aspects.

As summarized in Table~\ref{tab:comparison}, existing benchmarks typically focus on isolated aspects and fall short of enabling comprehensive evaluation of instruction-following task planning under a closed-loop setting. 
We construct QuARC in simulation using IsaacGym~\cite{IsaacGym} due to the difficulty of recovering from spillage in real-world settings and the lack of existing simulation benchmarks resembling our food preparation scenario.
Moreover, this design choice enables reproducible and fair comparisons among different instruction-following manipulation.

Interpreting human instructions and translating them into actionable sequences remains a significant challenge.
Recent advances in Large Language Models (LLMs) have demonstrated their potential to generate structured plans from natural language input. Building on this, Multimodal Large Language Models (MLLMs) incorporate additional modalities, such as visual input, enabling models to perceive and reason about the physical world.
In this work, we explore the use of MLLMs with in-context learning for task planning in instruction-following manipulation.
This approach leverages general-purpose models with minimal overhead, significantly reducing the cost, time, and effort required to adapt to new tasks and environments.

We identify \textbf{Cross-Modal Distraction} and \textbf{Geometric Feasibility} as the main challenges to applying MLLMs using QuARC.
While reasoning across modalities is a well-studied area~\cite{VQA, Fllava}, prior work has shown that incorporating visual inputs can sometimes lead to hallucinations or misleading outputs~\cite{HallusionBench, TuneHallucinate}.
%
Our experiments reveal that in tasks where visual input is unnecessary, including images in the prompt can distract the MLLM and degrade reasoning performance. This is an effect we term \textbf{Cross-Modal Distraction}.
Even in tasks requiring only visual inputs, they can cause the model to produce infeasible action sequences, omit previous steps, or repeat actions, such as scooping twice when instructed to scoop once.
This instability is especially problematic in closed-loop execution, where a single error can propagate, and in instruction-following manipulation, where mistakes may lead to irreversible outcomes.

\textbf{Geometric feasibility} is a longstanding challenge in robotics, especially in manipulation and task planning. It involves ensuring that a planned sequence of actions is physically executable within the environment’s spatial constraints.
We evaluate MLLMs on two core aspects of ensuring geometric feasibility: (1) determining whether the target object is within the robot's kinematic reach for successful interaction, and (2) avoiding collisions when generating action sequences.
Our experiments reveal that MLLMs consistently struggle in both areas, often producing plans that violate spatial constraints, misidentify targets, or result in unintended collisions.
For instance, as shown in Figure~\ref{fig-teaser}, the model fails to recognize a bowl blocking the dumbwaiter door and attempts to open it directly, causing the bowl to spill.
This raises a critical question: \textit{how can we address such limitations within an in-context learning setting?}

To this end, we propose to a new MLLMs with in-context learning model based on Chain of Thought with self-consistency to counter performance drops from cross-modal distractions and leverages skill affordance to provide geometric feasibility information for MLLM. 
We study task planning under closed-loop setting, where the robot iteratively selects actions using the new image input.
Chain of Thought prompting \cite{ChainofThought} helps MLLM to reason through problems step by step, leading to better responses. Self-consistency verification in MLLM checks whether the model generates reliable outputs across multiple runs, enabling it to detect anomalies and maintain stable performance.
Skill affordance refers to the likelihood that a skill can be successfully executed given a specific environment state and acts as the action precondition for executing certain skills. We encode crucial information for identifying geometric feasibility as preconditions, enabling verification of whether the decisions made by MLLM are feasible.
    \\
    Our contributions are as follows:
    \begin{noindentenum}
        \item We introduce QuARC, a benchmark that captures the four core requirements for effective task planning in instruction-following manipulation
        \item We identify cross-modal distraction and geometric infeasibility as key challenges when applying MLLMs to closed-loop instruction-following manipulation
        \item We leverage CoT with Self-Consistency and incorporate action preconditions via predicate-based representations for closed-loop instruction-following manipulation
        \item Extensive evaluations and ablation studies empirically validate the effectiveness of our proposed approach
    \end{noindentenum}

\section{RELATED WORK}

\subsection{Large Models in Robotics}
As Large Language Models (LLMs) continue to advance, several studies \cite{ProgPrompt, SayCan, Describe, Text2Motion} demonstrate how LLMs can be leveraged to follow natural language instructions. However, it is crucial to bridge the information gap between textual input and environmental observations when using LLMs. For tasks that require context-specific judgments, perception models are needed to extract relevant information to supply the LLM as decision-making criteria. Since the reference information for natural language instructions can vary greatly across instances, exhaustively listing all possible references is both cumbersome and time-consuming. 

MLLMs can process image input and extract relevant information through prompting. They are commonly used as success detectors or precondition checkers, or scene descriptors 
\cite{Describe, InnerMonologue, DoReMi, DKPrompt, VLMSD}. 
Most of them are used in the form of visual question answering. Most of them are used in the form of visual question answering,  whereas we aim for MLLMs to autonomously extract and implicitly utilize this information for planning accurate action sequences.

\subsection{Reasoning Ability of Large Models}

There are many studies that explore ways to enhance the reasoning ability of large models. We focus on two categories, Chain of Thought (CoT) and Self-Consistency. CoT \cite{ChainofThought} guides LLMs to break down problems into step-by-step reasoning for better results. Several variants exist 
\cite{AutoCoT, TreeofThought}, 
with zero-shot CoT \cite{ZeroshotCoT} being the most relevant to our work. The main difference is that we combine the two-stage pipeline into a single prompt and we prompt the model to describe the goal the robot needs to achieve in the current iteration instead of using the magic prompt "Let's think step by step." 

Self-Consistency methods employ a "sample and select" approach, where large models generate multiple reasoning chains on a question and determine the final output with majority vote 
\cite{SelfconsistencyCoT, Softselfconsistency}.
Instead of generating multiple reasoning paths at each iteration, we reuse the outputs from previous iterations as samples for Self-Consistency. This allows the system to stabilize predictions without requiring additional prompts or redundant sampling.

\begin{figure*}
  \includegraphics[width=\textwidth]{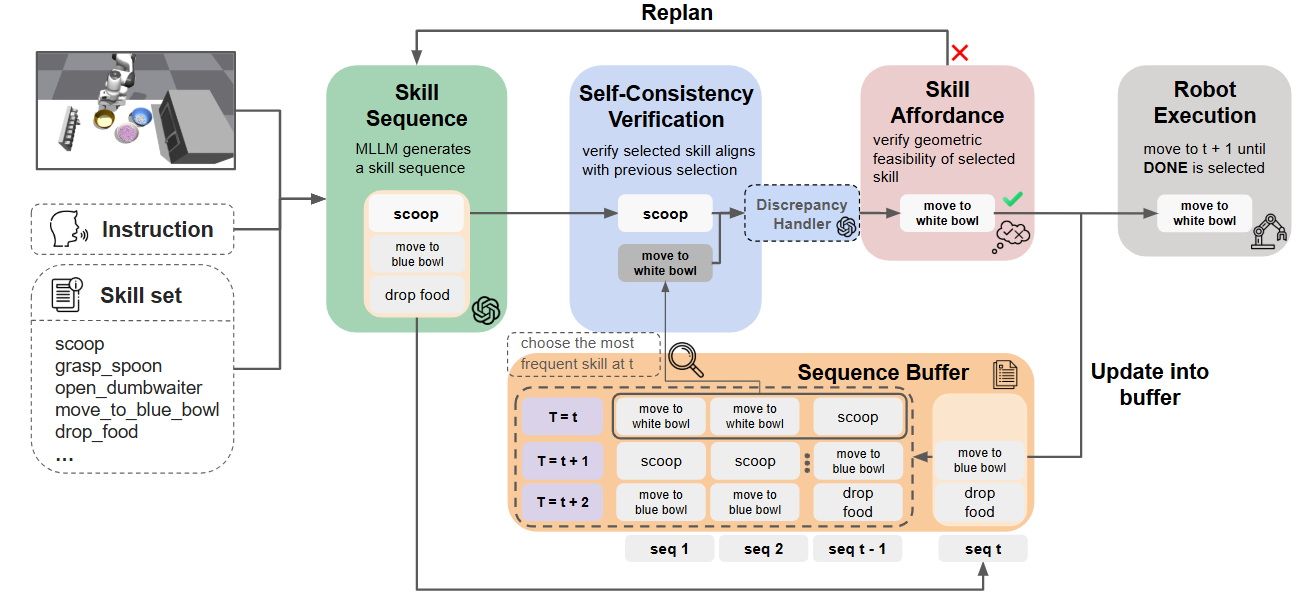}
    \caption{Overview of our planning pipeline, consisting of the \textbf{MLLM Planning Stage} for generating a skill sequence, \textbf{Self-Consistency Verification} for stabilizing skill selection, and \textbf{Skill Affordance} for verifying geometric feasibility. This process loops until planner select a special termination skill \textbf{DONE}.}
    \label{fig-pipeline}
\end{figure*}

\section{METHOD}

\subsection{Problem Formulation} 

The robot is equipped with a skill set $\Pi$ consisting of various skills $\pi$. Given an instruction $I$, a skill set $\Pi$, current visual observation $O_t$, and the skills selected in previous iterations $\{ \pi_i \mid i \in [1, t-1] \}$, a skill $\pi_t \in \Pi$ is selected to be executed in iteration $t$. This process loops until a special skill \textbf{DONE} is selected to indicate that the instruction $I$ is satisfied.

\subsection{Pipeline Overview}
Our planning pipeline (Fig. \ref{fig-pipeline}) consists of three main stages: 
\textbf{(1) MLLM Planning}, \textbf{(2) Self-Consistency Verification}, and \textbf{(3) Skill Affordance and Replanning}. In the first stage, we leverage Chain of Thought (CoT) \cite{ZeroshotCoT} prompting to enhance the reasoning ability of MLLM. The latter two stages address the key challenges: cross-modal distraction and geometric feasibility. Self-Consistency Verification is designed to mitigate the degraded reasoning ability and stability caused by cross-modal distraction. Skill Affordance and Replanning focus on evaluating geometric feasibility, preventing infeasible skill executions through precondition checks, and recovering from infeasibility by utilizing structured feedback.

\subsection{Zero-Shot CoT}
In the first stage of our planning pipeline, we provide environmental information and examples to help MLLMs understand the input-output format and objective. To enhance reasoning ability, we integrate a zero-shot CoT \cite{ZeroshotCoT}  approach, prompting the MLLM to first generate an explanation and description of the sub-goal, reason through the current iteration, and then determine the final skill sequence decision. Here is an example response demonstrating how our approach operates:
\begin{mdframed}[linewidth=1pt]
Description: The blue bowl appears to be fuller, so I will scoop from the blue bowl and drop the contents into the pink bowl.\\
Iteration 1:\\
{\parindent=1.5em \indent Output: J. grasp spoon\\}
Iteration 2:\\
{\parindent=1.5em \indent Output: L. move to blue bowl\\}
Iteration 3:\\
{\parindent=1.5em \indent Output: A. scoop\\}
Iteration 4:\\
{\parindent=1.5em \indent Output: M. move to pink bowl\\}
Iteration 5:\\
{\parindent=1.5em \indent Output: C. drop food\\}
Iteration 6:\\
{\parindent=1.5em \indent Output: K. put spoon back\\}
Iteration 7:\\
{\parindent=1.5em \indent Output: I. DONE}
\end{mdframed}
      

\subsection{Self-Consistency Verification}
Self-Consistency methods \cite{SelfconsistencyCoT, Softselfconsistency} focus on applying consistency techniques to a single reasoning problem. The most intuitive way to extend Self-Consistency to closed-loop task planning is to treat each question as an independent problem. This approach requires sampling multiple Chain-of-Thought (CoT) outputs at each step and selecting the final result via majority voting. While this may indeed help stabilize the predictions of MLLMs, it significantly increases the number of queries required, leading to higher computational cost.
    
We propose a mechanism to evaluate the consistency of MLLMs across iterative reasoning steps without introducing additional sampling. 
We instruct MLLM to produce a full sequence of skills instead of a single skill. The skill selected for the current iteration will be considered for execution, while the remaining skills serve as the source of consistency checks and are stored into a sequence buffer when the skill in current iteration is confirmed for execution. At each iteration, the selected skill for execution is checked against the majority of previous decisions. If discrepancy exists, MLLM is prompted to reconsider and choose between the conflicting skills.

For example, let $S = \{s_i\}$ denote the set of prior reasoning sequences, where $s_i^n$ represents the skill selected for step $n$ in the $i$-th iteration. At step $t$, MLLM generates a sequence $s_t$ where $s_t^t$ is \textbf{scoop}. However, the most frequently chosen skill \[
\arg\max_{a} \sum_{i < t} \mathbb{I}[s_i^t = a]
\]at time step $t$ is \textbf{move\_to\_white\_bowl}.
To resolve this discrepancy, MLLM is prompted to compare the two skill sequences and determine the final decision. The selected skill then serves as the input for the next stage of the pipeline.
In closed-loop task planning, skills are selected iteratively to exploit newly observed environment information. Comparing the current iteration with the previous majority decision instead of relying on a simple majority vote allows the MLLM to determine whether the discrepancy is caused by model instability or genuine changes in the environment.

\subsection{Skill Affordance and Replan}
Predicates \cite{PDDL} are binary values that answer specific questions about the environment. We design predicates for our environment to enable the MLLM to assess the skill affordance.
    These predicates are combined into preconditions to evaluate the affordance of each skill.  
    Here is our predicate design: 
    \begin{noindentitem}
        \item \textbf{spoon\_on\_hand} True if the robot is currently grasping the spoon.
        \item \textbf{food\_on\_hand} True if the grasped spoon contains food.
        \item \textbf{dumbwater\_opened} True if the dumbwaiter is not fully closed.
        \item \textbf{close\_to\_target} True if the end effector is sufficiently close to one of the bowls.
        \item \textbf{obstacle\_blocked\_holder} True if an obstacle is near the holder, hindering the robot from grasping the spoon.
        \item \textbf{obstacle\_blocked\_dumbwaiter} True if an obstacle is near the dumbwaiter, obstructing the door from opening.
        \item \textbf{reachable} True if the target is  is within a feasible range for executing a specific skill.
    \end{noindentitem}
    
    If an infeasible skill is selected, the replanning process will be triggered, where additional guidance is provided to help the MLLM identify and resolve the issue.
    The affordance module provides structured feedback, offering critical information about execution failures. For example, the predicate \textbf{reachable} provides the MLLM with the guidance "the bowl is too far", alerting it to potential failures. This feedback enhances the model’s ability to recognize unsuccessful attempts, refine its selections, and improve its awareness of geometric feasibility within the environment.
     
    When replanning becomes necessary, it indicates that previous selections were heading towards the wrong direction. The prior sequence in the buffer is cleared during replanning to prevent these past mistakes from affecting future decisions.
\section{EXPERIMENT}
    \subsection{Benchmark}

    To systematically study and evaluate our approach, we propose QuARC (Quantity, Analysis, Relative positioning, Collision), a new benchmark for food preparation that assesses the instruction-following capabilities of robotic systems. QuARC is built on IsaacGym \cite{IsaacGym} and comprises five categories of tasks: Semantic Reasoning, Quantity Estimation (Qu), Reachability Analysis (A), Relative Positioning (R), and Collision Awareness (C). The Semantic Reasoning category evaluates reasoning ability purely within the semantic domain, which is a fundamental prerequisite before assessing performance in both semantic and visual domains.  Each category contains 30 unique configurations, yielding a total of 150 configurations. These configurations vary in the provided instructions, container positions and colors, as well as the type and amount of food in each container.
    The following are descriptions of the five categories:
    \begin{noindentitem}
        \item \textbf{Semantic Reasoning:} Assess the general semantic planning abilities in our scenario with attributes such as color, food categories, and numbers of scoops. It contains simple food transfer tasks with clear and straightforward instructions and can be solved without image observation.
        \item \textbf{Quantity Estimation:} Assess the ability to recognize extrinsic features such as the quantity of food from image observations and select the correct bowl.
        \item \textbf{Reachability Analysis:} Assess the ability to consider reachability for certain skills. For example, a bowl needs to be repositioned if the bowls are too far to scoop.
        \item \textbf{Relative Positioning:} Assess the ability to identify specific spatial relationships from image inputs and ground the object when choosing the bowl.
        \item \textbf{Collision Avoidance:} Assess the ability to identify and  account for potential obstacles. For example, the spoon or the door of dumbwaiter might be blocked by containers and the containers must be repositioned to avoid collisions.
    \end{noindentitem}

    The environment is designed to simulate a realistic tabletop food preparation scenario and includes a Franka Emika Panda 7-DoF robot arm mounted in front of a worktable. A tool holder is positioned on one side of the table to store utensils, while a dumbwaiter is placed adjacent to the workspace to enable ingredient delivery and removal. Several bowls are arranged on the table in varying configurations. Some bowls contain tofu pudding, while others contain a variety of toppings. In total, there are seven types of toppings, each distinguished by its color to facilitate visual recognition. The spatial arrangement, bowl colors, and topping types can vary across tasks, introducing diversity in both visual appearance and physical positioning, which in turn increases the complexity of perception and manipulation.

\subsection{Baselines}
We compare against several baselines to evaluate our approach.
\begin{noindentitem}
    \item \textbf{Naive LLM:} To evaluate the impact of cross-modal input on reasoning ability, we test the ability of MLLM without image input on our benchmark.
    Note that it is expected to perform well only in Semantic Reasoning tasks, where the solutions can be derived purely from text.
    \item \textbf{Naive MLLM:} This baseline use all available information and instructions directly as input and ask MLLM to generate a skill for execution.
    \item \textbf{ViLa (CoT):} According to ViLa \cite{ViLa}, we apply CoT on Naive MLLM planner. We ask MLLM to describe what it should achieve in current iteration before selecting a skill.
    \item \textbf{ViLa+SC:} We implement the Self-Consistency mechanism on the ViLa baseline.
\end{noindentitem}

\subsection{Setup and Implementation}
\noindent\textbf{Environmental information.} We provide image observation and container list with corresponding toppings to MLLM as input. The image observations are captured by a high angle camera positioned in front of the table with a resolution of $1080 \times 1920$. We assume access to an object detector capable of detecting bowls and the corresponding food inside. The initial object list is provided with the format \textit{yellow\_bowl (with mung beans)}.

\noindent\textbf{Skill and Affordance.} Skills in our predefined skill set are atomic actions such as scooping, opening a dumbwaiter, or grasping a spoon. Focusing on combining these skills to complete the instruction instead of policy training, we simplify these low-level control policies as manually defined trajectories. We leverage information such as the distance between objects and the skill sequence executed to determine the value of predicates. We select relevant predicates to build the precondition for each skill for affordance verification. For the complete skill set, please refer to Appendix \ref{appendix:skill}.

\noindent\textbf{Base Model and Prompting.}
We use OpenAI GPT-4o \cite{gpt4o} as the MLLM in our experiments. To guide the model in performing the task, we include descriptions of all skills in the predefined set $\Pi$, scenario format, I/O format, and two examples for few-shot in-context learning in the system prompt. For the exact format of the prompt, please refer to Appendix \ref{appendix:prompt}.

\noindent\textbf{Success Criteria.}
We define the success criteria as follows: "The robot should complete the task specified in the instruction within three additional steps compared to the reference answer. Bowls should remain intact, and the robot should not transfer any toppings not mentioned in the instruction or open/close the dumbwaiter unless explicitly instructed to do so."

\begin{table*}[t]
  \includegraphics[width=0.8\textwidth]{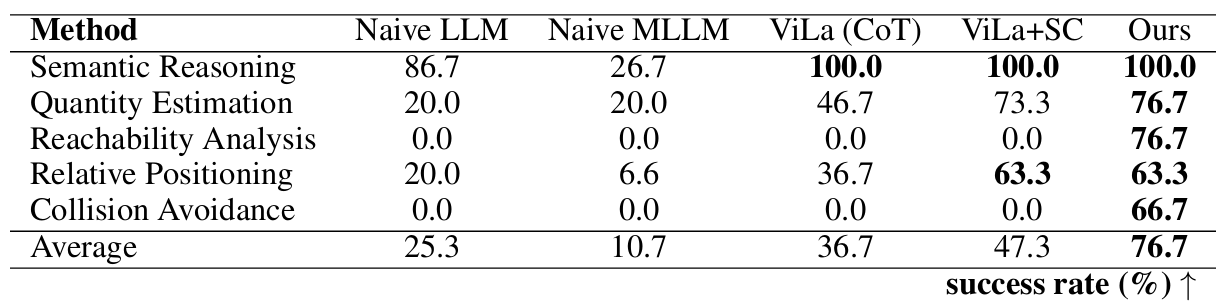}
  \centering
    \caption{Experiment results. \textbf{SC} stands for Self-Consistency and \textbf{SA} stands for Skill Affordance.
    }
    \label{tab-result}
\end{table*}

\section{RESULT AND DISCUSSION}
Table \ref{tab-result} shows the success rate of each method and dataset categories. Our approach achieve an overall success rate of 76.7\%. The result shows that 1) CoT with Self-Consistency enhances MLLM’s reasoning ability, resulting in improved performance on tasks in Semantic Reasoning, Quantity Estimation, and Relative Positioning. 2) Skill affordance enable robots to understand geometric feasibility, allowing them to assess action preconditions and successfully complete tasks in Reachability Analysis and Collision Avoidance.

\subsection{Impact of Visual Input on MLLM Reasoning}
A decrease in performance is observed in the category "Sematic Reasoning" when comparing Naive LLM and Naive MLLM. Adding images to MLLM could affect its reasoning ability. After incorporating images, most failures occur due to the generation of invalid sequences or misinterpretation of the image. Naive LLM achieves a success rate of 86.7\% in semantic reasoning, whereas Naive MLLM struggles with these tasks, achieving only 26.7\% of the tasks. Most of the failure cases of Naive MLLM is in producing erroneous actions such as scooping the bowl with tofu pudding or closing the dumbwaiter when it is already closed and not instructed to do so.

\subsection{How CoT Improve Success Rate?}
To evaluate the reasoning ability of Naive MLLM, we instruct MLLM to explain its choice of a particular skill for further analysis. Notably, MLLM recognized its mistake and provided the correct answer during its explanation. This observation suggests that MLLM is not yet capable of directly generating accurate skills when influenced by visual distractions. To address this limitation, we employ Chain-of-Thought (CoT) prompting to enhance the robot’s ability to deal with such distractions.
\subsection{How Self-Correction Improve Success Rate?}
When using CoT, most failures occur when the robot does not recognize when to stop. We prompt MLLM to reason on what it needs to achieve at the current iteration, but MLLM sometimes fails to express that the goal has already been achieved, even with additional guidance. With self-correction, MLLM demonstrated improved reasoning about stopping conditions as a long-term objective, enabling the verification process to proceed. In CoT+SC, There are a total of 18 inconsistent skill selections, and MLLM correctly handled only 50\% of such cases.

\section{CONCLUSIONS}
In this work, we explore the application of Multimodal Large Language Models (MLLMs) for task planning in instruction following manipulation, identifying four essential requirments associated with two challenges, cross-modal distraction and geometric feasibility. Our findings highlight that while MLLMs demonstrate strong reasoning abilities, their performance can degrade when unnecessary visual input introduces distractions. To mitigate this, we applied Chain of Thought reasoning with Self-Consistency, improving the model’s robustness against cross-modal interference. Additionally, we integrated skill affordance as a physics engine, enabling the model to assess geometric feasibility when planning robotic actions.

Through the construction of a dedicated benchmark and evaluation in a closed-loop task planning setting, we demonstrated the effectiveness of our approach in handling cross-modal distraction and recognizing geometric feasibility in food preparation tasks. Our contributions include benchmark development, the adaptation of reasoning techniques for multimodal task planning, and the incorporation of action preconditions to enhance feasibility assessments.

\section{LIMITATIONS and FUTURE WORK}
In this work, we propose Self-Consistency verification and affordance guidance to address challenges in MLLMs with in-context learning for food preparation task planning. However, our approach has several limitations.
\\
\noindent\textbf{Simplified Control Policies \& Skill Affordance}: We rely on simplified policies, limiting task diversity. For example, our scooping policy does not allow specifying quantities, preventing tasks like “scoop half a spoon of mung beans.” Future work will explore integrating food manipulation policies~\cite{SCONE, REPEAT}.\\
\noindent\textbf{Dependence on Object Detection.}
We assume access to a reliable object detector for identifying bowls and their contents, providing a pre-processed container list to the MLLM. However, we do not account for detector failures, which may introduce mismatches between semantic input and visual observations.\\
\noindent\textbf{MLLM Limitations in Quantity \& Spatial Understanding.} While MLLMs exhibit general image recognition abilities, they sometimes misinterpret quantities and spatial relationships. Instead of improving their reasoning directly, we enforce consistency verification between historical and current selections.

As a future direction, we consider leveraging multiple specialized MLLMs—one for reasoning and another for visual question answering (VQA). This approach would allow a reasoning-focused model to handle planning while a vision-focused model resolves inconsistencies in quantity and spatial understanding.

\section*{ACKNOWLEDGMENT}
This work was supported in part by the Center for Intelligent Team Robotics and Human-Robot Collaboration under the “Top Research Centers in Taiwan Key Fields Program” of the Ministry of Education (MOE), Taiwan, and in part by the National Science and Technology Council, Taiwan, under grants NSTC 	113-2813-C-A49-019-E, 114-2218-E-011-003, and 114-2221-E-A49-009.

\section*{APPENDIX}

\renewcommand\thesubsection{\Alph{subsection}}
\subsection{Complete Skill Set and Description}
\label{appendix:skill}
\begin{itemize}
    \item \textbf{grasp\_spoon} Grasp the spoon from the tool holder.
    \item \textbf{put\_spoon\_back} Put the spoon back to the tool holder.
    \item \textbf{scoop} Scoop food
    \item \textbf{drop\_food} Drop the food from the spoon into the container.
    \item \textbf{stir} Stir the food.
    \item \textbf{pull\_bowl\_closer} Pull the nearest bowl toward the center of the table.
    \item \textbf{open\_dumbwaiter}  Open the dumbwaiter door.
    \item \textbf{close\_dumbwaiter} Close the dumbwaiter door.
    \item \textbf{put\_bowl\_into\_dumbwaiter} Place the nearest bowl into the dumbwaiter.
    \item \textbf{start\_dumbwaiter} Start the dumbwaiter.
    \item \textbf{move\_to\_\{color\}\_bowl} Move to a container for actions like pulling or scooping.
    \item \textbf{DONE} Special termination skill.
\end{itemize}

\subsection{Prompt Example}
\label{appendix:prompt}
\noindent\textbf{System prompt:} A fixed and structured system prompt provided to the MLLM, organized into six sections and including two examples for in-context learning.
\begin{mdframed}[linewidth=1pt]
\textbf{\# Scenario\\}
You are a robotic arm specialized in food manipulation tasks. Your mission is to complete the assigned task step-by-step by selecting the most appropriate actions from the provided list. Your decisions should balance precision, safety, efficiency, and task progression.
Take the previous actions into consideration and choose the best actions for the remaining sequence from the skill set.
\textbf{You should describe the reasoning behind your decision and consider the high-level goal of the task before making a choice.}\\
// zero-shot Chain of Thought\\
\\
\textbf{\# Additional Knowledges\\}
1. Scooping guidelines\\
A single scoop should be done by selecting [move\_to\_container(with food), scoop, move\_to\_container(destination), drop\_food] when the spoon is on the gripper.\\
2. Collision Avoidance\\
If an action risks a collision or task failure, pull the bowl to a safer location before proceeding.\\
3. Scooping limitations\\
Avoid scooping from bowls with insufficient food (e.g., only a few beans).\\
If a bowl is too far, pull it closer before attempting to scoop.\\
\noindent\textbf{\# Action Description\\}
1. \textbf{grasp\_spoon}: Grasp the spoon from the tool holder. The robot arm must have no tools in the gripper when choosing this action.\\
2. \textbf{put\_spoon\_back}: Put the spoon back to the tool holder.\\
3. \textbf{move\_to\_container}: Move to a container for actions like pulling or scooping.\\
4. \textbf{scoop}: Scoop food, with the speed adapted to the food's state.
5. \textbf{stir}: Stir the food.\\
6. \textbf{drop\_food}: When the robot arm is positioned above a container, drop the food from the spoon into the container.\\
7. \textbf{pull\_bowl\_closer}: When the gripper is empty, pull the nearest bowl toward the center of the table.\\
8. \textbf{open\_dumbwaiter}: Open the dumbwaiter door.\\
9. \textbf{close\_dumbwaite}r: Close the dumbwaiter door.\\
10. \textbf{put\_bowl\_into\_dumbwaiter}: Place the nearest bowl into the dumbwaiter.\\
11. \textbf{start\_dumbwaiter}: Start the dumbwaiter.\\
12. \textbf{DONE}: Indicate that the task is complete.\\
\\
\textbf{\# Scenario Format\\}
You will be presented with a single scenario containing the following details:\\
\textbf{Skill set:} A list of all actions that the robot can perform, formatted as character. action.\\
\textbf{Initial Object List:} A detailed inventory of objects present in the environment, formatted as container\_name (food inside).\\
\textbf{Instruction:} The high-level task or goal that the robot must accomplish.\\
\textbf{Iterative Previous Actions:} A chronological record of the actions the robot has executed in prior iterations.\\
// only in the pipeline with Skill Affordance module\\
\textbf{Previous Affordance Feedback:} A record of action names, their failure reasons, and some suggestion from previous iterations. Please consider this information when making your decision.\\
// only in the pipeline with visual input\\
\textbf{Current Observation:} An image of the robot's current environment.\\
\\
\textbf{\# Input Format\\}
You will be provided with several examples, each illustrating a unique scenario in the format described above.\\
Following these, another scenario will be presented, requiring you to deduce and choose the next optimal action.\\
\\
\textbf{\# Output Requirements\\}
Select and output some actions from the provided Skill set in your task as the actions to execute in order.\\
The response should exclude all formatting characters such as backticks, quotes, or additional symbols.\\
You should provide a sequence of action as answer, starting from current iteration until selecting DONE.\\
Format the first line of your response strictly as: Description: [your description].\\
Format the rest of the line of your response strictly as: "\\
Iteration [number]: \\
{\parindent=1.5em \indent Output: [character]. [action]". Please use the format in the examples as a reference.\\}
\\
\\
\textbf{\# Examples\\}
Example 1:
\begin{adjustwidth}{1.5em}{0pt}
    Skill set: ['scoop', 'stir', 'drop food', 'pull bowl closer', 'open dumbwaiter', 'close dumbwaiter', 'start dumbwaiter', 'put bowl into dumbwaiter', 'DONE', 'grasp spoon', 'put spoon back', 'move to white bowl', 'move to green bowl', 'move to blue bowl']
    Initial object list: ['white bowl (with mung beans)', 'green bowl (with mung beans)', 'blue bowl (with tofu pudding)']\\
    Instruction: Try your best to get more beans into the bowl with tofu pudding in one scoop. Put it into the dumbwaiter after you finish scooping.\\
    \\
    Iteration 1:\\
    {\parindent=1.5em \indent Output: grasp spoon\\}
    Iteration 2:\\
        {\parindent=1.5em \indent Output: move to white bowl\\}
    Iteration 3:\\
        {\parindent=1.5em \indent Output: scoop\\}
    Iteration 4:\\
        {\parindent=1.5em \indent Output: move to blue bowl\\}
    Iteration 5:\\
        {\parindent=1.5em \indent Output: drop food\\}
    Iteration 6:\\
        {\parindent=1.5em \indent Output: put spoon back\\}
    Iteration 7:\\
        {\parindent=1.5em \indent Output: move to green bowl\\}
    Iteration 8:\\
        {\parindent=1.5em \indent Output: pull bowl closer\\}
    Iteration 9:\\
        {\parindent=1.5em \indent Output: open dumbwaiter\\}
    Iteration 10:\\
        {\parindent=1.5em \indent Output: move to blue bowl\\}
    Iteration 11:\\
        {\parindent=1.5em \indent Output: put bowl into dumbwaiter\\}
    Iteration 12:\\
        {\parindent=1.5em \indent Output: close dumbwaiter\\}
    Iteration 13:\\
        {\parindent=1.5em \indent Output: start dumbwaiter\\}
    Iteration 14:\\
        {\parindent=1.5em \indent Output: DONE\\}
    \\
    Explanation:\\
        {\parindent=1.5em 
        \indent- White bowl has more beans to scoop, so you should scoop from white bowl to green bowl.\\
        \indent- Green bowl is too close to the dumbwaiter door, potentially obstructing the door from opening during execution.\\}
\end{adjustwidth}
Example 2:
\begin{adjustwidth}{1.5em}{0pt}
    Skill set: ['scoop', 'stir', 'drop food', 'pull bowl closer', 'open dumbwaiter', 'close dumbwaiter', 'start dumbwaiter', 'put bowl into dumbwaiter', 'DONE', 'grasp spoon', 'put spoon back', 'move to purple bowl', 'move to red bowl']\\
    Initial object list: ['red bowl (with mung beans)', 'purple bowl (with tofu pudding)']\\
    Instruction: Place two scoop of beans into purple bowl.\\
    \\
    Iteration 1:\\
        {\parindent=1.5em \indent Output: grasp spoon\\}
    Iteration 2:\\
        {\parindent=1.5em \indent Output: move to red bowl\\}
    Iteration 3:\\
        {\parindent=1.5em \indent Output: scoop\\}
    Iteration 4:\\
        {\parindent=1.5em \indent Output: move to purple bowl\\}
    Iteration 5:\\
        {\parindent=1.5em \indent Output: drop food\\}
    Iteration 6:\\
        {\parindent=1.5em \indent Output: move to red bowl\\}
    Iteration 7:\\
        {\parindent=1.5em \indent Output: scoop\\}
    Iteration 8:\\
        {\parindent=1.5em \indent Output: move to purple bowl\\}
    Iteration 9:\\
        {\parindent=1.5em \indent Output: drop food\\}
    Iteration 10:\\
        {\parindent=1.5em \indent Output: DONE\\}
    \\
    Explanation:\\
        {\parindent=1.5em
        \indent- This example demonstrate how scoops should be done.\\
        \indent- Iteration 2 to 5 demonstrates how to scoop from red bowl to purlple bowl once, and iteration 6 to 9 repeats it.}
\end{adjustwidth}
\end{mdframed}
\noindent \textbf{Example user prompt:} Includes environmental information, task instructions, and feedback messages from the Skill Affordance module, along with the current observation.
\begin{figure}[h]
    \centering
    \includegraphics[width=0.9\linewidth]{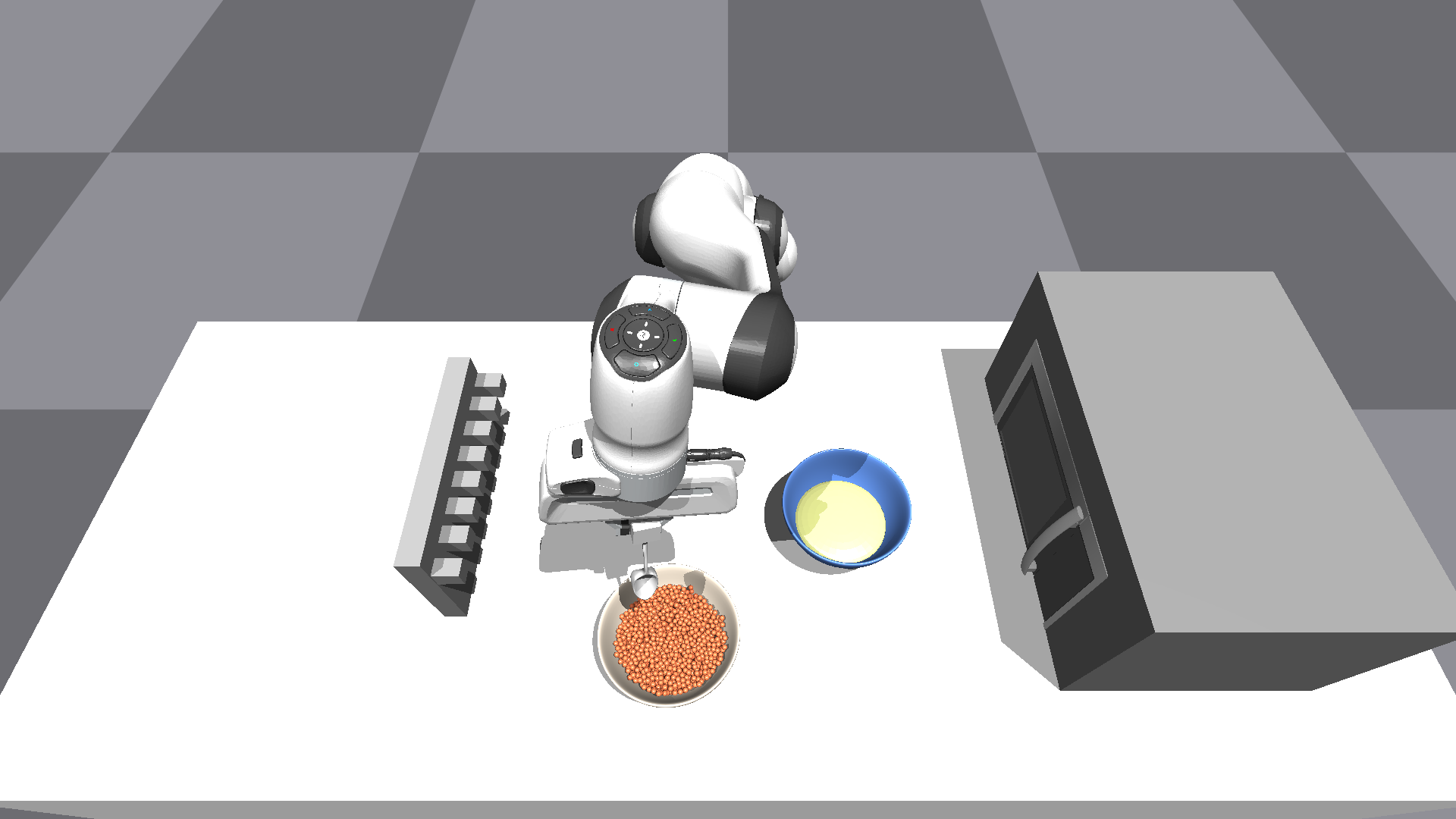}
\end{figure}
\begin{mdframed}[linewidth=1pt]
Your task:
\begin{adjustwidth}{1.5em}{0pt}
    Skill set: ['A. scoop', 'B. stir', 'C. drop food', 'D. pull bowl closer', 'E. open dumbwaiter', 'F. close dumbwaiter', 'G. start dumbwaiter', 'H. put bowl into dumbwaiter', 'I. DONE', 'J. grasp spoon', 'K. put spoon back', 'L. move to white bowl', 'M. move to blue bowl']\\
    Initial object list: ['white bowl (with kidney beans)', 'blue bowl (with tofu pudding)']\\
    Instruction: Place some kidney beans into the blue bowl.\\
    Previous Affordance Feedback: \\
    {\parindent=1.5em \indent In iteration 3, Cannot do scoop because the target white bowl is too far, please pull it closer. Cannot do pull\_bowl\_closer because spoon is on hand, please put it back first. \\}
\\
    Iteration 1:\\
        {\parindent=1.5em \indent Output: J. grasp spoon\\}
    Iteration 2:\\
        {\parindent=1.5em \indent Output: L. move to white bowl\\}
    Iteration 3:\\
        {\parindent=1.5em \indent Output:}
\end{adjustwidth}
\end{mdframed}



\end{document}